\documentclass{article}
\usepackage{spconf,amsmath,graphicx}
\usepackage{multirow}
\usepackage{comment}
\usepackage{mdframed}
\usepackage{tipa}

\title{T5lephone: Bridging Speech and Text Self-supervised Models for Spoken Language Understanding via Phoneme level T5}
\name{Chan-Jan, Hsu$^{1 *}$\thanks{$^*$ These authors contributed equally.}, Ho-Lam, Chung$^{2 *}$, Hung-Yi, Lee$^{2 }$, Yu, Tsao$^{3 }$}
\address{$^{1 }$MediaTek Research  
$^{2 }$National Taiwan University, Taiwan 
$^{3 }$Academia Sinica, Taiwan }
\begin{document}

\ninept
\maketitle
\begin{abstract}
In Spoken language understanding (SLU), a natural solution is concatenating pre-trained speech models (e.g. HuBERT) and pretrained
language models (PLM, e.g. T5). 
Most previous works
use pretrained language models with subword-based tokenization.
However, the granularity of input units affects the alignment of
speech model outputs and language model inputs, and PLM with
character-based tokenization is underexplored.
In this work, we conduct extensive studies on how PLMs with
different tokenization strategies affect spoken language understanding task including spoken question answering (SQA) and
speech translation (ST). 

We further extend the idea to create T5lephone\footnote{pronounced as telephone}, a variant of T5 that is pretrained using phonemicized text. We initialize T5lephone with  existing PLMs to pretrain it using relatively lightweight computational resources. We reached state-of-the-art on NMSQA, and the T5lephone model exceeds T5 with other types of units on end-to-end SQA and ST. Our code is publicly available.\footnote{https://github.com/Splend1d/T5lephone}
\end{abstract}
%

\section{Introduction}
\label{sec:intro}

Spoken language understanding (SLU) aims to not only decipher but also comprehend audio signals. A well-trained SLU model could be applied to solve tasks such as spoken question answering (SQA) and speech translation (ST). Following the success of self-supervised text pretraining \cite{devlin2018bert, lewis2019bart, raffel2020exploring, he2020deberta}
, self-supervised speech pretraining \cite{DBLP:conf/nips/BaevskiZMA20,DBLP:journals/taslp/HsuBTLSM21,DBLP:journals/corr/abs-2110-13900} 
aims to learn strong speech representation for downstream tasks. Despite reaching near-perfect performance on speech intent classification and speech keyword spotting on SUPERB \cite{yang2021superb}, these models' performance on ST using a randomly initialized transformer decoder \cite{tsai2022superb} is not competitive with works that incorporate knowledge from text \cite{li2020multilingual, conneau2022xtreme}. Pretrained language models (PLM) are also frequently present in previous works solving SQA \cite{chuang2019speechbert,chung2020splat,lin2022dual}. We thus conclude from previous works that incorporating textual knowledge in the system is desired. 

There are two main ways to include pretrained language model in the SLU system. Cascaded approaches utilize raw text as the anchor to link the speech representations and the word representations \cite{shon2022slue,huang2022mtl}. The more recent end-to-end approach seeks to reduce error prorogation by using speech representations directly as the input of pretrained language models \cite{li2020multilingual, lin2022dual}, and it is logical to do so since language models are cross-disciplinary learners \cite{kao2021bert}. However, most works use a subword-based tokenization level language model in their system, such as sentencepiece or byte-pair encoding (bpe). Other tokenization strategies, such as character-based tokenization are often overlooked. For example, the SLUE benchmark \cite{shon2022slue} uses DeBERTa \cite{he2020deberta} in cascaded speech understanding, DUAL \cite{lin2022dual} uses Longformer \cite{beltagy2020longformer} in end-to-end speech question answering, and \cite{li2020multilingual} uses mBART \cite{liu2020multilingual} in the speech translation task. These three pretrained language models all incoporate subword-based tokenization.

The input unit granularity of the PLM is potentially important in semantic speech tasks for both cascaded and end-to-end methods. For cascaded systems, ASR error degrades performance, but an incorrectly recognized word may have characters that resemble the gold label. Therefore, the character error rate (CER) of the ASR results might be lower than the word error rate (WER), and using PLMs with character-level inputs is theoretically beneficial.
 For end-to-end systems utilizing a pretrained language model, mitigating the degree of mismatch between speech representations and the original pretraining text data may also boost system performance. Since self-supervised speech representations have been found to resemble phoneme sequences after clustering and reduction \cite{baevski2021unsupervised} when using 128 clusters, they are much more similar to character level inputs than to subword level inputs.

In this work, we conduct an extensive study on how self-supervised PLMs with different input granularity affect SQA/ST performance, by inferring on datasets such as NMSQA \cite{lin2022dual} and Covost2 \cite{wang2021covost}. In particular, we compared T5/mT5 with ByT5, which has similar pretraining settings. 
We then further extend the idea to create T5lephone, a variant of T5 that takes phonemicized text as input. T5lephone is realized by self-supervised second-phase pretraining \cite{DBLP:journals/corr/abs-2004-10964} using phonemicized text from Wikipedia, with the model being initialized from mT5/ByT5. We devised a novel way to re-represent the phonemicized text to maximize the transferable knowledge from original text pretraining of ByT5 to our phoneme pretraining.  We reached state-of-the-art and +12\% performance gain on previous cascaded NMSQA results\cite{lin2022dual} while using fewer parameters. The performance of our T5lephone model also exceeds previous methods on end-to-end NMSQA and ST.
\begin{figure*}
    \centering
    \includegraphics{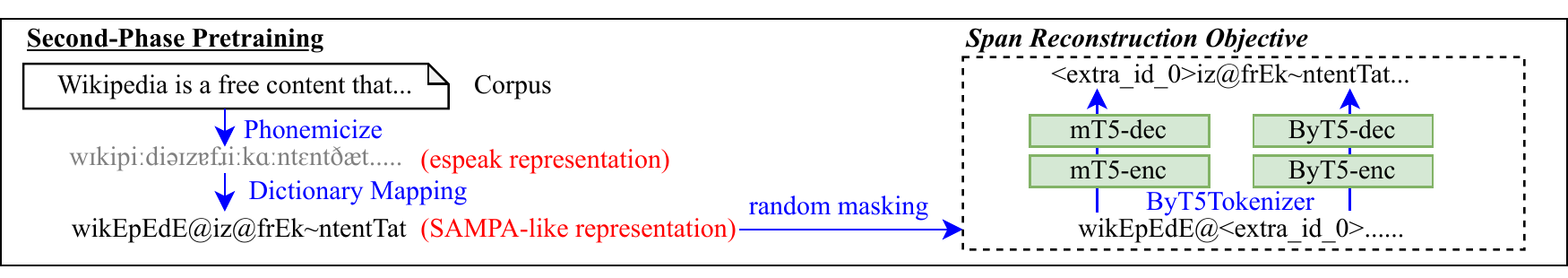}
    \caption{
    Pipeline for our second-phase pretraining initialized with different variants of T5 models. The text corpus is preprocessed using the espeak phonemizer followed by a dictionary mapping. The T5 model is trained using the span reconstruction objective with the phonemicized inputs. 
    }
    \label{fig:pretraining}
\end{figure*}

\section{Proposed Method}
\label{sec:method}

\subsection{Pretrained Modules}
All of our experimented downstream tasks require a speech model followed by a language model. The speech model is responsible for extracting speech information, which is either vector representations or ASR outputs. The information is then forwarded into the language model for task-specific training.  

For the speech model, we use self-supervised models such as wav2vec2.0 and HuBERT to extract representations. Since self-supervised learning alone is not sufficient to produce ASR results, we use the checkpoint that is finetuned on Librispeech in such cases.  

For the pretrained language models, we choose the generative model T5 and its variants (mT5/ByT5) to allow for a wide range of tasks and comparisons, noting that a generative model may also perform extractive tasks, by using the encoder only which similar to the methods in EncT5\cite{liu2021enct5}. mT5 is a multilingual version of T5, and ByT5 is also multilingual while using byte-level inputs. ByT5 has shown strong robustness against noised textual inputs over mT5\cite{xue2022byt5}, which might extend beyond artificially generated noise to ASR noise in cascaded systems. Our T5lephone is a model adapted from mT5/ByT5 which we will explain in the following subsection.

\subsection{T5lephone - Phoneme Input T5}

 To create a model that takes in phoneme sequence as input, we conduct second-stage pretraining to the variants of T5. That is, we use mT5/ByT5 as initialization and train the model with the original span reconstruction objective\cite{xue2022byt5} using phonemicized inputs from the wiki text corpus, as shown in Figure~\ref{fig:pretraining}. The resulting model is named T5lephone. 
 \subsubsection{Pretraining from Byt5}
 There are multiple standardized ways to represent text data as phoneme sequences \cite{baevski2021unsupervised, tang2021general,tang2021improving}, and each of these methods can be supported by ByT5 tokenizer, since the ByT5 tokenizer (that is just reading the sequence byte by byte) supports universal character set decoding. We use a two-stage approach as shown in Figure \ref{fig:pretraining} to represent phoneme sequences. The text sequence is first converted to espeak using a phonemizer, and then converted to SAMPA-like representations using a dictionary. The second conversion step is necessary because some characters in the espeak phoneme set such as "\texttt{\ae}" is decoded into two bytes and unnecessarily extend the sequence length. On the other hand, the SAMPA-like chart contains all ASCII characters which are one byte only. The mapping characters are inspired by the SAMPA chart\footnote{https://en.wikipedia.org/wiki/SAMPA\_chart}, based on the IPA. The full mapping table can be found in our code. The mapped sequence has the additional benefit of being closer to the original text sequence (measured by CER). For preprocessing, we eliminate all spaces between words in resemblance to the speech model output not having word boundaries. 
\begin{table*}
\centering
\small
\begin{tabular}{|c|c|c|c|cc|cc|cc|cc|}
\hline

\multicolumn{12}{|c|}{\textbf{Cascaded Spoken Question Answering}}\\
\hline
\multirow{2}{*}{\textbf{ASR Model}} &\multirow{2}{*}{\textbf{PLM}}&\multirow{2}{*}{\textbf{tokenization}} & \multirow{2}{*}{\#Params} & \multicolumn{2}{c|}{\textbf{text dev}} & \multicolumn{2}{c|}{\textbf{dev}}  & \multicolumn{2}{c|}{\textbf{test-SQuAD}} & \multicolumn{2}{c|}{\textbf{test-OOD}} \\
\cline{5-12}
 & & &&EM & F1 & AOS & FF1 & AOS & FF1 & AOS & FF1\\
\hline
w2v2-ft-960h & longformer-base &subword& 148M & 85.0  & 91.9 & 47.7 & 58.6 & 50.1 &  62.5 &  43.9 & 53.6\\

w2v2-ft-960h & deberta-large  & subword&405M & 87.9  & 93.9 & 42.0 & 52.2 &  48.6 & 61.5 &  33.1 & 42.5 \\
w2v2-ft-960h & T5-small &subword&61M& 78.9  & 86.1 &49.3&55.5&45.3&53.2&35.2&45.1 \\

w2v2-ft-960h & T5-base& subword&222M& 83.0 & 89.9 & 55.6 & 62.8 & 66.6 & 73.9 & 37.9 & 44.3 \\
w2v2-ft-960h & T5-large &subword&770M& 84.2 & 91.8 & \textbf{65.2} & \textbf{70.0} & 66.5 & 72.5 & 52.4 & 56.3 \\
\hline
w2v2-ft-960h & ByT5-small&byte&299M &78.4 &83.9&60.0&64.7& 69.9 & 74.1 & 53.9 & 57.7  \\

w2v2-ft-960h & ByT5-base &byte&581M& 80.6 & 87.0 & 64.0 & 68.8 & 68.6 & 73.5 & \textbf{60.9} & \textbf{66.1} \\
w2v2-ft-960h & \textbf{ByT5lephone-small}&byte&299M &76.7 &83.7&59.2&64.4 &\textbf{70.5} & \textbf{75.5} & 58.3 & 63.3 \\
\hline
\hline
w2v2-ft-10min & longformer-base &subword&148M & 85.0  & 91.9 &  44.8 &55.3 & 50.7 &  63.4 &  38.4 & 47.3\\

w2v2-ft-10min & deberta-large &subword&405M & 87.9  & 93.9 & 38.3 & 49.1 &  47.5 & 60.3 &  28.7 & 37.6 \\
w2v2-ft-10min & T5-small&subword &61M& 78.9  & 86.1 &44.7 &  51.3&44.6&51.5&29.0&35.1 \\

w2v2-ft-10min & T5-base &subword&222M& 83.0 & 89.9 & 52.8 & 60.4 & 58.2 & 66.7 & 37.6 & 43.7 \\
w2v2-ft-10min & T5-large &subword&770M& 84.2 & 91.8 & 57.6 & 62.8 & 58.8 & 65.5 & 48.4 & 52.4  \\
\hline 
w2v2-ft-10min & ByT5-small&byte &299M&78.4 &83.9&55.4&60.6& 62.8 & 67.7 & 41.3 & 45.4  \\

w2v2-ft-10min & ByT5-base &byte&581M& 80.6 & 87.0 & \textbf{59.7} & \textbf{65.3} & 65.2 & 69.7 & 48.4 & \textbf{53.3} \\

w2v2-ft-10min & \textbf{ByT5lephone-small} &byte&299M&76.7 &83.7&55.0&60.8& \textbf{70.1} & \textbf{76.3} & \textbf{48.6} & 53.2  \\
\hline
\hline
sew-d-tiny-ft-100h & ByT5-small&byte &299M& 78.4 &83.9 & \textbf{49.4} & 54.5 & 38.4 & 42.8 & \textbf{40.2} & \textbf{45.1} \\
sew-d-tiny-ft-100h & \textbf{ByT5lephone-small}&byte &299M& 76.7 &83.7 & \textbf{49.4} & \textbf{55.1} & \textbf{51.0} & \textbf{57.0} & 35.8 & 40.5 \\
\hline
\multicolumn{12}{|c|}{Training (and testing) with Phonemicized text (and Phoneme ASR)}\\
\hline
-- & \textbf{ByT5lephone-small} &byte&299M& 52.4 & 67.8 & -- & -- & -- & -- & -- & -- \\
\hline
\end{tabular}
\caption{
Cascaded Spoken Question Answering results. EM stands for the exact match metric, AOS stands for area overlapping Score, and FF1 is the frame level F1 score. For each speech model section, the horizontal line separates models that use subword-based tokenization (e.g. sentencepiece) from models that use character-based tokenization. For each ASR model, the best score is marked in \textbf{bold}.
}
\label{tab:CascadedSQA}
\end{table*}
 \subsubsection{Pretraining from mT5}
 
 Sentencepiece is used in mT5. It is vulnerable to text data that is foreign to the model or misspelled words in general, so we cannot apply it to phoneme sequences. Therefore, we still use the ByT5 tokenizer to convert the phonemicized text to ids as shown in Figure \ref{fig:pretraining}. We then forward the input id sequence to the model using the vocabulary embedding from the initialized model type for self-supervised training. This is similar to training from scratch, but there is no downside of initializing the model with the trained weights\cite{kao2021bert}. By doing so, we show generalizability of our pretraining method to subword PLMs.

 \subsubsection{Pretraining Details}
Our span reconstruction objective uses a masking ratio equal to 15\% and an average span equal to 20 drawn from a poisson distribution. We train the model for 200000 steps with a batch size of 128 and the learning rate equal to 3e-4. To give a clear indication of the initialized model type, we prepend its prefix to the name T5lephone in the experimental session. For cascaded SQA and ST, we found out that using a the 20000-step checkpoint is suffice and training further does not increase score.

\subsection{Speech Language Understanding Tasks}
We tested our T5lephone model on three tasks, including cascaded SQA, end-to-end SQA, and speech to text translation. For SQA, we use NMSQA, which is a publicly available spoken question answering dataset \cite{lin2022dual}. The training and development sets are derived from SQuAD \cite{rajpurkar2016squad}, and the speech is generated with Amazon Poly. The test set is derived from SQuAD, NEWSQA \cite{trischler2016newsqa} and QuAC \cite{choi2018quac}, and the speech is read by human. Since NEWSQA and QuAC are out-of-domain samples, they are labeled as OOD and often reported alongside test-SQuAD. The correct answer is labeled by a time span, and the metrics are AOS (Area Overlapping Score) and FF1 (Frame F1). For speech to text translation, Covost2 \cite{wang2021covost} is a publicly available multilingual speech translation dataset. In our experiments, we only experiment with En$->$De portion of the dataset. 
\subsubsection{Testing on Cascaded SQA}
For cascaded SQA, we trained the text models on text SQuAD, and tested on ASR text of NMSQA. The ASR text is generated by wav2vec2 finetuned on LibriSpeech \cite{panayotov2015librispeech}. For extractive models, the training method follow \cite{devlin2018bert}, while for generative models, we reframe the problem to generative question answering following \cite{raffel2020exploring}.  Both methods can produce answers in the form of a timespan, which is required for the calculation of the FF1/AOS metrics. For more details we refer the reader to our code.

For generated answers that cannot be extracted from text input, the score is 0 for both AOS and FF1, but this rarely happens as all of the training text data is from an extractive QA dataset, so the model learns to only output word sequences that are present in the input.

\subsubsection{Finetuning on end-to-end SQA}
\label{sec:e2eft}
\begin{table}
\centering
\small
\begin{tabular}{|c|c|c|c|}
\hline

\multicolumn{4}{|c|}{\textbf{SQA ASR Error}}\\
\hline
\textbf{ASR Model}  & \textbf{dev}  & \textbf{test-SQuAD} & \textbf{test-OOD} \\
\hline
w2v2-ft-960h  & 11.53 & 11.28  & 17.29  \\
w2v2-ft-10min &  16.05 &  16.54 & 21.64\\
sew-d-tiny-ft-100h &26.11 & 32.75 & 36.62\\
\hline

\end{tabular}
\caption{
ASR word error rate for different ASR models on the NMSQA dataset for the cascaded spoken question answering task.
}
\label{tab:WER}
\end{table}
For end-to-end SQA, we generally follow the process of \cite{lin2022dual}, choosing HuBERT as the speech model to benefit from its quantized reconstruction pertaining, as well as allowing comparison with previous works. First, we extract representations from HuBERT and conduct K-means clustering to obtain an integer sequence of the representations. Then, the output is reduced by merging unique consecutive tokens into a single token, and the sequence length is typically 3000 to 4000. Extractive QA is then trained using the condensed HuBERT sequence as input ids. Since there isn't any intermediate text output, generative methods are no longer possible in this setting, so we use the encoder only for T5 models to make it extractive. Unlike \cite{lin2022dual}, we do not discard sequences over the max length limit. Instead, we slice the context into several segments to concatenate it with the question and force the model to find the correct segment and the correct span\footnote{This is identical to how text SQuAD in implemented in huggingface}. We are thus able to lower our model length constraint to 1024 to include ByT5. Calculating AOS/FF1 is straightforward since each HuBERT token directly correlates to a fixed timespan. We train the model for 10 epochs with a learning rate equal to 3e-5, and choose the model according to dev set score.

\subsubsection{Finetuning on Speech to Text Translation}
The speech translation pipeline follows \cite{liu2020multilingual}, and the system is trained end-to-end with no frozen parameters. First, the speech representation is generated via a self-supervised speech model. Then the speech representation is passed through an adapter and a linear layer, to shorten the sequence as well as change the dimensions of the embeddings. The adaptor is a stack of three 1D convolutional layers with stride 2, and results in a sequence downsampling rate of 8. The pretrained language decoder attends to the embedding sequence of the linear layer output and learns to generate target text. We train the model for 20 epochs with a learning rate equal to 4e-5. Unlike SQA, in ST, it has already been reported in \cite{liu2020multilingual} that end-to-end methods beat cascaded methods so we do not experiment with cascaded ST.
\begin{figure}
    \centering
    \includegraphics{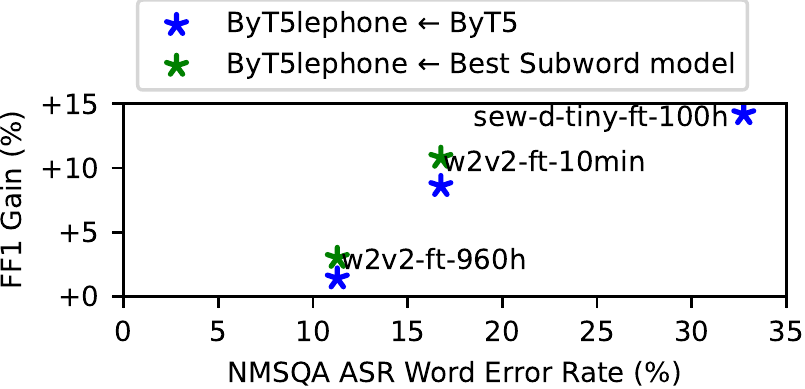}
    \caption{Peformance gains after swapping to our ByT5lephone. The gains positively correlates with the ASR error rates. Here the performance gains is measured on the test-SQuAD portion of the dataset.}
    \label{fig:diff}
\end{figure}
\section{Experimental Results}
\label{sec:exp}

\subsection{Cascaded SQA}
The experimental results are shown in Table \ref{tab:CascadedSQA}. Bold numbers in the table represent the best model for each setting (same speech model). The first column corresponds to the result evaluated on the dev set of text SQuAD, to evaluate the model performance when ASR error is not present. The single horizontal line in between each ASR model setting separates the subword-level models with the byte-level models. While subword models such as longformer and deberta-large are the top performers on text SQuAD, when testing on NMSQA with ASR error, byte-level models significantly outperform subword-level language models. This is clearly evident in the results using the best ASR model w2v2-ft-960h. In this scenario, ByT5-small can even outperform deberta-large, which is the best in text SQuAD and has more parameters. Furthermore, our second-phase pretrained ByT5lephone further outperforms ByT5-small on test-SQuAD and test-OOD by 1\% to 5\%, and reaching state-of-the art on NMSQA test-SQuAD. 

To see how ASR error affects performance, we also tested with less robust ASR systems, which are smaller (sew-d-tiny-ft-ls100h\cite{https://doi.org/10.48550/arxiv.2109.06870}\footnote{Squeezed and Efficient Wav2vec finetuned on 100 hours of LibriSpeech}) and/or trained on less data (w2v2-ft-10min\footnote{Wav2vec2.0 finetuned on 10 minutes of LibriSpeech}). As shown in Figure \ref{fig:diff}, the performance gap between the two models positively correlates with the word error rate of the speech model. As ASR error increases from 11.28 to 32.75 on test-SQuAD, the improvement of ByT5lephone over ByT5 shoots from 1\% to 15 \%. This shows that pretraining on phoneme sequences (ByT5lephone-small) induces better ASR error denoising capabilities. The ASR word error rates are shown in Table \ref{tab:WER} for reference. 

We acknowledge that there is also the possibility of using an ASR system that directly produces phonemes, then forward it into ByT5lephone to solve NMSQA. However, training and testing phonemicized text SQuAD already significantly underperforms on the dev set (Table \ref{tab:CascadedSQA}), therefore we conclude that swapping to a phoneme ASR would not further improve NMSQA in this setting.

\begin{table}
\centering
\small
\begin{tabular}{|c|c|cc|cc|}
\hline

\multicolumn{6}{|c|}{\textbf{End-to-end Spoken Question Answering}}\\
\hline 
\textbf{Text Model} 
&
\multirow{2}{*}{\textbf{Len}} 
&  
\multicolumn{2}{c|}{\textbf{test-SQuAD}} 
& 
\multicolumn{2}{c|}{\textbf{test-OOD}} \\
\cline{3-6}
 (enc. only) && AOS & FF1 & AOS & FF1 \\
\hline
longformer\cite{lin2022dual} & 4096 &     49.1  &  55.9 &-&-\\
 longformer & 4096 &     46.0  &  53.9 & 32.2 & 36.9 \\ 

\hline 
\hline
 longformer & 1024 &    40.0 &  46.0 & 22.8 & 26.4 \\ 
 ByT5-small& 1024  &   48.4 &  54.9 & 27.1 & 31.0  \\ 
 \textbf{ByT5lephone-small} & 1024  & \textbf{53.3} &  \textbf{61.1} & \textbf{32.3} & \textbf{37.3}  \\ 

\hline
\end{tabular}
\caption{
End-to-end Spoken Question Answering results. "Len" is the max sequence length of the model. AOS stands for area overlapping Score, and FF1 is the frame level F1 score. Longformer-1024 is the same model as longformer-4096 but with the maximum sequence length capped at 1024.
}
\label{tab:TextlessSQA}
\end{table}

\begin{table}
\centering
\small
\begin{tabular}{|c|c|c|}
\hline

\multicolumn{3}{|c|}{\textbf{Speech to Text Translation En$->$De}}\\
\hline
 \textbf{Text Model} & \multirow{2}{*}{\textbf{\#Params}} & \multirow{2}{*}{\textbf{BLEU}} \\
(dec. only)&&\\

\hline
 mT5-small & 153M & 26.8  \\
\textbf{mT5lephone-small} & 153M & \textbf{27.2}   \\
\hline
 ByT5-small&  82M & 27.9  \\
 \textbf{ByT5lephone-small} & 82M & \textbf{28.1}   \\

\hline
\end{tabular}
\caption{
Speech to Text Translation Results. BLEU score is calculated with the sacreBLEU package.
}
\label{tab:ST}
\end{table}

\subsection{End-to-end SQA}
The end-to-end SQA results are shown in Table \ref{tab:TextlessSQA}. We use the dev set to select the model, and report the test scores. From Table \ref{tab:TextlessSQA}, it can be seen that our ByT5lephone outperforms ByT5 and longformer-4096, improving end-to-end NMSQA by 7\% on test-SQuAD. This improvement is further amplified by the fact that observing that the max sequence length of the model makes a huge difference. As shown by comparing longformer-4096 with longformer-1024 (same model but max sequence length capped to 1024), longformer-1024 underperforms by 6\% to 10\%. When taken maximum sequence into account, ByT5lephone-small outperforms longformer-1024 by 9\% to 16\%. Finally, although our experiment with longformer-4096 is slightly lower than the reported results in \cite{lin2022dual}, our method includes entries with sequences longer than 4096 by methods described in section \ref{sec:e2eft}, so our setting is also more challenging.
\begin{figure}
    \centering
    \includegraphics[width=7.5cm]{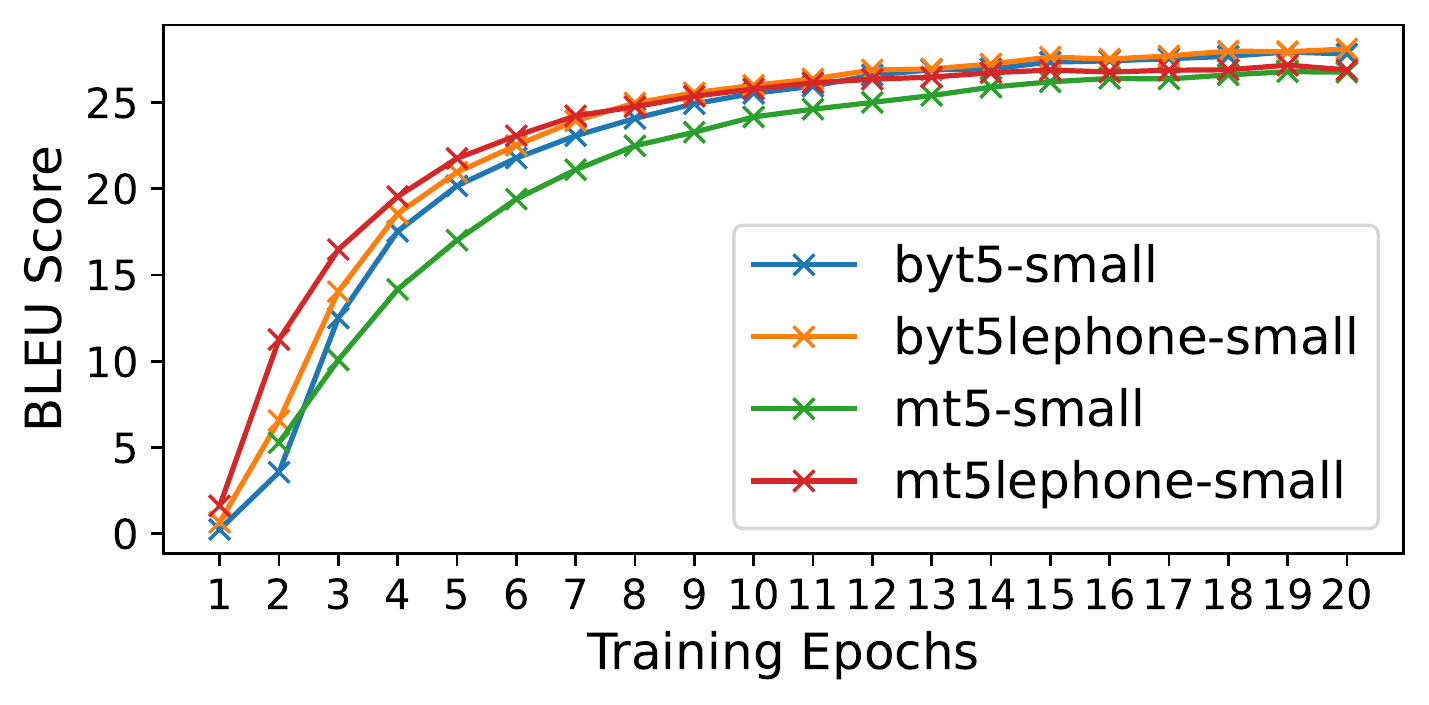}
    \caption{Training Curve for the ST Task. T5lephone variants converges faster than their respective base model.  }
    \label{fig:BLEU}
\end{figure}

Overall, the results show that our second-phase finetuning is effective. Further, the findings on the effect on sequence length on performance motivate future work on very long PLMs in various settings. 

\subsection{Speech To Text Translation}
The ST results are shown in Table \ref{tab:ST}. We observe that ByT5 outperforms mT5 with fewer parameters on the Covost2 En→De dataset, showing that changing to byte-level inputs has a positive effect even when only the decoder of the PLM is used. Our T5lephone models also constantly outperform their base model. These results are comparable with previous works\cite{liu2020multilingual} using mBART as the PLM. We also plot the BLEU scores of the intermediate checkpoints in Figure \ref{fig:BLEU} to show that T5lephone converges much faster than its respective base model, proving the effectiveness of our second-phase pretraining. It can also be seen that ByT5 converges faster than mT5, thus the superiority of ByT5 is consistent with our findings in cascaded SQA.

\section{Conclusion}
\label{sec:Conclustion}
In this work, we first justify the use of byte-level models over subword-level model in SLU. We then extended this idea and conducted second-phase pretraining of PLMs on phonemicized text, which reaches SOTA on cascaded SQA, while also gaining performance on end-to-end SQA and speech to text translation.

\bibliographystyle{IEEEbib}
\bibliography{strings,refs} 

\begin{thebibliography}{10}

\bibitem{devlin2018bert}
Jacob Devlin, Ming-Wei Chang, Kenton Lee, and Kristina Toutanova,
\newblock ``Bert: Pre-training of deep bidirectional transformers for language
  understanding,''
\newblock {\em arXiv preprint arXiv:1810.04805}, 2018.

\bibitem{lewis2019bart}
Mike Lewis, Yinhan Liu, Naman Goyal, Marjan Ghazvininejad, Abdelrahman Mohamed,
  Omer Levy, Ves Stoyanov, and Luke Zettlemoyer,
\newblock ``Bart: Denoising sequence-to-sequence pre-training for natural
  language generation, translation, and comprehension,''
\newblock {\em arXiv preprint arXiv:1910.13461}, 2019.

\bibitem{raffel2020exploring}
Colin Raffel, Noam Shazeer, Adam Roberts, Katherine Lee, Sharan Narang, Michael
  Matena, Yanqi Zhou, Wei Li, Peter~J Liu, et~al.,
\newblock ``Exploring the limits of transfer learning with a unified
  text-to-text transformer.,''
\newblock {\em J. Mach. Learn. Res.}, vol. 21, no. 140, pp. 1--67, 2020.

\bibitem{he2020deberta}
Pengcheng He, Xiaodong Liu, Jianfeng Gao, and Weizhu Chen,
\newblock ``Deberta: Decoding-enhanced bert with disentangled attention,''
\newblock {\em arXiv preprint arXiv:2006.03654}, 2020.

\bibitem{yang2021superb}
Shu-wen Yang, Po-Han Chi, Yung-Sung Chuang, Cheng-I~Jeff Lai, Kushal Lakhotia,
  Yist~Y Lin, Andy~T Liu, Jiatong Shi, Xuankai Chang, Guan-Ting Lin, et~al.,
\newblock ``Superb: Speech processing universal performance benchmark,''
\newblock {\em arXiv preprint arXiv:2105.01051}, 2021.

\bibitem{tsai2022superb}
Hsiang-Sheng Tsai, Heng-Jui Chang, Wen-Chin Huang, Zili Huang, Kushal Lakhotia,
  Shu-wen Yang, Shuyan Dong, Andy~T Liu, Cheng-I~Jeff Lai, Jiatong Shi, et~al.,
\newblock ``Superb-sg: Enhanced speech processing universal performance
  benchmark for semantic and generative capabilities,''
\newblock {\em arXiv preprint arXiv:2203.06849}, 2022.

\bibitem{li2020multilingual}
Xian Li, Changhan Wang, Yun Tang, Chau Tran, Yuqing Tang, Juan Pino, Alexei
  Baevski, Alexis Conneau, and Michael Auli,
\newblock ``Multilingual speech translation with efficient finetuning of
  pretrained models,''
\newblock {\em arXiv preprint arXiv:2010.12829}, 2020.

\bibitem{conneau2022xtreme}
Alexis Conneau, Ankur Bapna, Yu~Zhang, Min Ma, Patrick von Platen, Anton
  Lozhkov, Colin Cherry, Ye~Jia, Clara Rivera, Mihir Kale, et~al.,
\newblock ``Xtreme-s: Evaluating cross-lingual speech representations,''
\newblock {\em arXiv preprint arXiv:2203.10752}, 2022.

\bibitem{chuang2019speechbert}
Yung-Sung Chuang, Chi-Liang Liu, and Hung-Yi Lee,
\newblock ``Speechbert: Cross-modal pre-trained language model for end-to-end
  spoken question answering,''
\newblock 2019.

\bibitem{chung2020splat}
Yu-An Chung, Chenguang Zhu, and Michael Zeng,
\newblock ``Splat: Speech-language joint pre-training for spoken language
  understanding,''
\newblock {\em arXiv preprint arXiv:2010.02295}, 2020.

\bibitem{lin2022dual}
Guan-Ting Lin, Yung-Sung Chuang, Ho-Lam Chung, Shu-wen Yang, Hsuan-Jui Chen,
  Shang-Wen Li, Abdelrahman Mohamed, Hung-yi Lee, and Lin-shan Lee,
\newblock ``Dual: Textless spoken question answering with speech discrete unit
  adaptive learning,''
\newblock {\em arXiv preprint arXiv:2203.04911}, 2022.

\bibitem{shon2022slue}
Suwon Shon, Ankita Pasad, Felix Wu, Pablo Brusco, Yoav Artzi, Karen Livescu,
  and Kyu~J Han,
\newblock ``Slue: New benchmark tasks for spoken language understanding
  evaluation on natural speech,''
\newblock in {\em ICASSP 2022-2022 IEEE International Conference on Acoustics,
  Speech and Signal Processing (ICASSP)}. IEEE, 2022, pp. 7927--7931.

\bibitem{huang2022mtl}
Zhiqi Huang, Milind Rao, Anirudh Raju, Zhe Zhang, Bach Bui, and Chul Lee,
\newblock ``Mtl-slt: Multi-task learning for spoken language tasks,''
\newblock in {\em Proceedings of the 4th Workshop on NLP for Conversational
  AI}, 2022, pp. 120--130.

\bibitem{kao2021bert}
Wei-Tsung Kao and Hung-Yi Lee,
\newblock ``Is bert a cross-disciplinary knowledge learner? a surprising
  finding of pre-trained models' transferability,''
\newblock {\em arXiv preprint arXiv:2103.07162}, 2021.

\bibitem{beltagy2020longformer}
Iz~Beltagy, Matthew~E Peters, and Arman Cohan,
\newblock ``Longformer: The long-document transformer,''
\newblock {\em arXiv preprint arXiv:2004.05150}, 2020.

\bibitem{liu2020multilingual}
Yinhan Liu, Jiatao Gu, Naman Goyal, Xian Li, Sergey Edunov, Marjan
  Ghazvininejad, Mike Lewis, and Luke Zettlemoyer,
\newblock ``Multilingual denoising pre-training for neural machine
  translation,''
\newblock {\em Transactions of the Association for Computational Linguistics},
  vol. 8, pp. 726--742, 2020.

\bibitem{baevski2021unsupervised}
Alexei Baevski, Wei-Ning Hsu, Alexis Conneau, and Michael Auli,
\newblock ``Unsupervised speech recognition,''
\newblock {\em Advances in Neural Information Processing Systems}, vol. 34, pp.
  27826--27839, 2021.

\bibitem{wang2021covost}
Changhan Wang, Anne Wu, Jiatao Gu, and Juan Pino,
\newblock ``Covost 2 and massively multilingual speech translation.,''
\newblock in {\em Interspeech}, 2021, pp. 2247--2251.

\bibitem{DBLP:journals/corr/abs-2004-10964}
Suchin Gururangan, Ana Marasovic, Swabha Swayamdipta, Kyle Lo, Iz~Beltagy, Doug
  Downey, and Noah~A. Smith,
\newblock ``Don't stop pretraining: Adapt language models to domains and
  tasks,''
\newblock {\em CoRR}, vol. abs/2004.10964, 2020.

\bibitem{liu2021enct5}
Frederick Liu, Siamak Shakeri, Hongkun Yu, and Jing Li,
\newblock ``Enct5: Fine-tuning t5 encoder for non-autoregressive tasks,''
\newblock {\em arXiv preprint arXiv:2110.08426}, 2021.

\bibitem{xue2022byt5}
Linting Xue, Aditya Barua, Noah Constant, Rami Al-Rfou, Sharan Narang, Mihir
  Kale, Adam Roberts, and Colin Raffel,
\newblock ``Byt5: Towards a token-free future with pre-trained byte-to-byte
  models,''
\newblock {\em Transactions of the Association for Computational Linguistics},
  vol. 10, pp. 291--306, 2022.

\bibitem{tang2021general}
Yun Tang, Juan Pino, Changhan Wang, Xutai Ma, and Dmitriy Genzel,
\newblock ``A general multi-task learning framework to leverage text data for
  speech to text tasks,''
\newblock in {\em ICASSP 2021-2021 IEEE International Conference on Acoustics,
  Speech and Signal Processing (ICASSP)}. IEEE, 2021, pp. 6209--6213.

\bibitem{tang2021improving}
Yun Tang, Juan Pino, Xian Li, Changhan Wang, and Dmitriy Genzel,
\newblock ``Improving speech translation by understanding and learning from the
  auxiliary text translation task,''
\newblock {\em arXiv preprint arXiv:2107.05782}, 2021.

\bibitem{rajpurkar2016squad}
Pranav Rajpurkar, Jian Zhang, Konstantin Lopyrev, and Percy Liang,
\newblock ``Squad: 100,000+ questions for machine comprehension of text,''
\newblock {\em arXiv preprint arXiv:1606.05250}, 2016.

\bibitem{trischler2016newsqa}
Adam Trischler, Tong Wang, Xingdi Yuan, Justin Harris, Alessandro Sordoni,
  Philip Bachman, and Kaheer Suleman,
\newblock ``Newsqa: A machine comprehension dataset,''
\newblock {\em arXiv preprint arXiv:1611.09830}, 2016.

\bibitem{choi2018quac}
Eunsol Choi, He~He, Mohit Iyyer, Mark Yatskar, Wen-tau Yih, Yejin Choi, Percy
  Liang, and Luke Zettlemoyer,
\newblock ``Quac: Question answering in context,''
\newblock {\em arXiv preprint arXiv:1808.07036}, 2018.

\bibitem{panayotov2015librispeech}
Vassil Panayotov, Guoguo Chen, Daniel Povey, and Sanjeev Khudanpur,
\newblock ``Librispeech: an asr corpus based on public domain audio books,''
\newblock in {\em 2015 IEEE international conference on acoustics, speech and
  signal processing (ICASSP)}. IEEE, 2015, pp. 5206--5210.

\bibitem{https://doi.org/10.48550/arxiv.2109.06870}
Felix Wu, Kwangyoun Kim, Jing Pan, Kyu Han, Kilian~Q. Weinberger, and Yoav
  Artzi,
\newblock ``Performance-efficiency trade-offs in unsupervised pre-training for
  speech recognition,'' 2021.

\end{thebibliography}
\end{document}